\documentclass{article}

\usepackage{arxiv}

\usepackage[utf8]{inputenc} 
\usepackage[T1]{fontenc}    
\usepackage{hyperref}       
\usepackage{url}            
\usepackage{booktabs}       
\usepackage{amsfonts}       
\usepackage{nicefrac}       
\usepackage{microtype}      
\usepackage{lipsum}
\usepackage{cite}
\usepackage{amsmath}
\usepackage{graphicx}
\usepackage{float}

\title{Classification of Multimodal Hate Speech -The Winning Solution of Hateful Memes Challenge}

\author{
  Xiayu Zhong\\
  Shijingshan District,Beijing,China\\
  \texttt{904666125@qq.com} \\
}

\begin{document}
\maketitle

\begin{abstract}
Hateful Memes is a new challenge set for multimodal classification, focusing on detecting hate speech in multimodal memes. Difficult examples are added to the dataset to make it hard to rely on unimodal signals, which means only multimodal models can succeed. According to Kiela\cite{kiela2020hateful},the state-of-the-art methods perform poorly compared to humans (64.73\% vs. 84.7\% accuracy) on Hateful Memes. I propose a new model that combined multimodal with rules, which achieve an first ranking of accuracy and AUROC of 86.8\% and 0.923 respectively. These rules are extracted from training set, and focus on improving the classification accuracy of difficult samples.
\end{abstract}


\section{Introduction}
\paragraph{The Hateful Memes competition hosted by Facebook.}The goal of this competition is to create an algorithm that identifies multimodal hate speech in internet memes, and this is a binary classification problem with multimodal input data consisting of the meme image itself and a string representing the text in the meme image. 

Internet memes are often harmless and sometimes hilarious. However, by using certain types of images, text, or combinations of each of these data modalities, the seemingly non-hateful meme becomes a multimodal type of hate speech, a hateful meme. At the massive scale of the internet, the task of detecting multimodal hate is both extremely important and particularly difficult, relying on just text or just images to determine whether or not a meme is hateful is insufficient. 

In this competition, model performance and leaderboard rankings were determined by the AUROC, and the accuracy scores of those submissions provided for additional information:
\begin{equation}
    AUROC = \int_{\infty}^{-\infty} \mbox{TPR}(T) \mbox{FPR}'(T) \, dT
\end{equation}
\begin{equation}
    Accuracy = \frac{1}{N} \sum_{i=0}^N I(y_i = \hat{y_i})
\end{equation}
The AUROC is calculated for each label in the submission and then averaged across the labels, which ranges from 0 to 1. Accuracy calculates the percentage of rows where the predicted class $\hat{y}$ in the submission matches the actual class, $y$ in the test set, the maximum is 1 and the minimum is 0. The goal of the competition is to maximize the AUROC.

\paragraph{My Contribution.}I proposed a classification algorithm that combined the multimodal with rules. First, I extracted classification rules from training set, and these rules could be used to generate pseudo-label or adjust classification probability. Using rules above, part of samples in test set can almost be determined, expending these samples to training set can greatly improve the metrics of model. I adjusted the predictions of each model with rules, then stack models to generate final prediction.

\paragraph{Overall competition results.}I achieved the first ranking with the highest AUROC and the highest accuracy in the phases 2.

\section{Related Work}
Self-supervised learning utilizes original data as its own source of supervision, which has been applied to many Computer Vision tasks, such as image colorization, solving jigsaw puzzles, inpainting, rotation prediction, and relative location prediction. Recently, pre-trained language models, such as BERT\cite{devlin2018bert}, RoBERTa\cite{liu2019roberta} and ALBERT\cite{lan2019albert}, have pushed great advances for NLP tasks. There are two keys to their success: effective pre-training tasks over large language corpus, and the use of Transformer for learning contextualized text representations.

There has been a surging interest in self-supervised learning for multimodal tasks, by pre-training on large-scale image/video and text pairs, then finetuning on downstream tasks. For example, VideoBERT\cite{sun2019videobert} and CBT\cite{sun2019learning} applied BERT to learn a joint distribution over video frame features and linguistic tokens from video-text pairs. ViLBERT\cite{lu2019vilbert} and LXMERT\cite{tan2019lxmert} introduced the two-stream architecture, where two Transformers are applied to images and text independently, which is fused by a third Transformer in a later stage. On the other hand, B2T2\cite{alberti2019fusion}, VisualBERT\cite{li2019visualbert}, Unicoder-VL\cite{li2020unicoder} and VL-BERT\cite{su2019vl} proposed the single-stream architecture, where a single Transformer is applied to both images and text. VLP\cite{zhou2020unified} applied pre-trained models to both image captioning and VQA\cite{antol2015vqa}. Furthermore, UNITER\cite{chen2020uniter} outperformed state-of-the-art models above over multiple V+L tasks by a significant margin. 12in1\cite{lu202012} found multi-task training could lead to significant gains over independent task training. 

Among the models above, UNITER and LXMERT were the best single-stream architecture and  the best two-stream architecture respectively. My attempt to improve the baselines provided by Facebook mainly refer to these two models.

\section{Dataset and Exploration}
\subsection{Dataset}
Memes pose an interesting multimodal fusion problem: Consider a sentence like "love the way you smell today" or "look how many people love you". Unimodally, these sentences are harmless, but combine them with an equally harmless image of a skunk or a tumbleweed, and suddenly they become mean.That is to say, memes are often subtle and while their true underlying meaning may be easy for humans to detect, they can be very challenging for AI systems.

The challenge sponsor constructed a dev and test set from 5\% and 10\% of the data respectively, and set aside the rest to serve as fine-tuning training data. This dataset totally contains more than 10k memes. The dev and test set were fully balanced, and were comprised of memes using the following percentages: 40\% multimodal hate, 10\% unimodal hate, 20\% benign text confounder, 20\% benign image confounder, 10\% random non-hateful. The objective of the task is, given an image and the pre-extracted text, to classify memes according to their hatefulness.

Dataset of this challenge in phase 2 is divided to five part: train, dev seen, test seen, dev unseen, test unseen. Two different phases share the same training set, but about 10\% of the labels are changed in phase 2. The dev seen and the test seen are datasets for phase 1 ,the dev unseen and test unseen are datasets for phase 2.

\subsection{Exploration and Rules Extraction}
After analyzing the merged set of training set and two dev sets, I found that there were a large percentage of memes that image or text appeared more than once, which accounted for about 30.7\% and 38.4\% respectively. If I defined the relationship of memes as the same image or the same text, these two relationship would not be mutually exclusive, while the proportion of independence samples with image and text appear just once is 45.4\%. The relationship between samples that established by the same image or the same text maybe very complicated. 

The meme images were clustered using a clustering algorithm based on perceptual hash, as there is a lot of text interference on the meme images. The meme texts were clustered by simple comparison on their strings. Through the clustering, the data in the training set were divided into 4 categories, which were "3-tuple", "2-tuple", "Unimodal hate" and others. The "3-tuple" consists of 3 memes, the first meme has the same image with the second meme, it also has the same text with the third meme, but the second meme and the third meme does not seem to be related. The "2-tuple" consists of 2 memes, they have the same image or the same text, and neither of them belongs to "3-tuple". The "Unimodal hate" is an image or text that it appear more than once in the training set, and the labels of all memes that contain this image or text are hateful. According to statistical analysis, the labels of “3-tuple” are 1, 0 and 0, where 1 represents hateful and 0 represents non-hateful, and most of the labels of “2-tuple” are 1 and 0. From the analysis above, I extract rules from training set:
\begin{itemize}
   \item  \textbf{rule 1.} For samples in "3-tuple", adjust the hateful probabilities to (1,0,0).
   \item  \textbf{rule 2.} For samples in "2-tuple", adjust the hateful probabilities to (1,0), where the sample has larger hateful probability adjusted to 1.
\end{itemize}
I implement rule 1 to get the label of "3-tuple" in test set, then merge these samples into training set, and implement rule 2 to adjust hateful probability of "2-tuple".

\section{Experiments and Results}
\begin{figure}[H]
  \centering
  \includegraphics[width=0.7\textwidth]{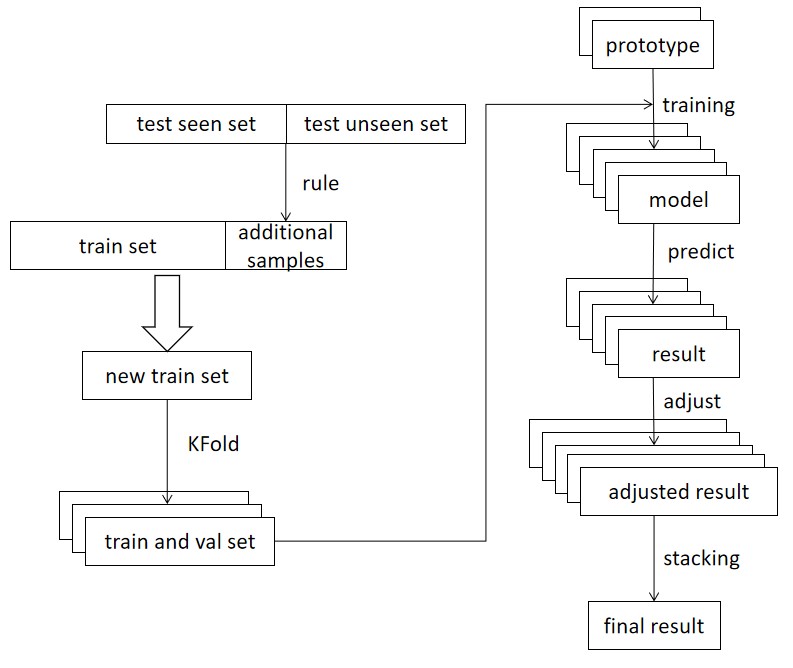}
  \caption{Process of my algorithm.}
  \label{fig:process}
\end{figure}

This section gives an overview of the selected models and the process to generate the final submissions. The process is shown in fig\ref{fig:process}.

\subsection{Pretrained Model Selection}
\label{subsec:selection}
Multimodal Modular Framework(MMF)\cite{singh2020mmf},which recommended by the organizer, is a modular framework for vision and language multimodal research from Facebook AI Research(FAIR), and it provides many multimodal pre-trained models of vision and language. The leaderboard of this competition shows that the models with AUROC scores above 0.7 are VisualBERT, VisualBERT COCO, ViLBERT, ViLBERT CC and MMBT-Region\cite{kiela2020hateful}. The AUROC scores on validation dataset (scores seen in table \ref{tab:baseline}) shows that VisualBERT outperform other algorithms mentioned above.

\begin{table}[H]
 \caption{scores of baselines provided by MMF}
  \centering
  \begin{tabular}{lll}
    \toprule
    Pretrained model key     & Model Key     & AUROC \\
    \midrule
    visual\_bert.finetuned.hateful\_memes.direct & visual\_bert &  0.7214\\
    visual\_bert.finetuned.hateful\_memes.from\_coco & visual\_bert &  0.7297\\
    vilbert.finetuned.hateful\_memes.direct & vilbert & 0.7135\\
    vilbert.finetuned.hateful\_memes.from\_cc\_original & vilbert & 0.7181\\
    mmbt.hateful\_memes.features & mmbt & 0.6893	\\
    lxmert & lxmert & 0.6880\\
    \bottomrule
  \end{tabular}
  \label{tab:baseline}
\end{table}

The classification results of different training rounds of each pre-trained model show that the best training rounds is around 3000, which means evaluate a pre-trained model on validation dataset takes about half an hour. So I had enough time to evaluate all finetuned and pre-trained models that MMF provided for hateful memes. These models are shown in table \ref{tab:models}, and the outperforming pre-trained models and their scores are listed in the table \ref{tab:scores}.

\begin{table}[H]
 \caption{Finetuned and pre-trained models provided by MMF}
  \centering
  \begin{tabular}{ll}
    \toprule
    Pretrained model     & Pretrained Key \\
    \midrule
    VisualBERT  & visual\_bert.finetuned.hateful\_memes.direct \\
    VisualBERT COCO  & visual\_bert.finetuned.hateful\_memes.from\_coco \\
    Masked COCO 100\%   & visual\_bert.pretrained.coco.full     \\
    Masked COCO 50\%   & visual\_bert.pretrained.coco.fifty\_pc     \\
    Masked COCO 10\%   & visual\_bert.pretrained.coco.ten\_pc     \\
    Masked COCO Train+Val   & visual\_bert.pretrained.coco.full\_train\_val   \\
    Masked VQA2 100\%   & visual\_bert.pretrained.vqa2.full \\
    Masked VQA2 50\%   & visual\_bert.pretrained.vqa2.fifty\_pc \\
    Masked VQA2 10\%   & visual\_bert.pretrained.vqa2.ten\_pc \\
    Masked VQA2 Train+Val   & visual\_bert.pretrained.VQA2.full\_train\_val   \\
    Masked CC 100\%   & visual\_bert.pretrained.cc.full     \\
    Masked CC 50\%   & visual\_bert.pretrained.cc.half     \\
    Masked CC 10\%(CC Small 100\%)   & visual\_bert.pretrained.cc.small     \\
    Masked CC Small 50\% (CC Small 50\%)   & visual\_bert.pretrained.cc.small\_fifty\_pc     \\
    Masked CC Small 10\% (CC Small 10\%)   & visual\_bert.pretrained.cc.small\_ten\_pc     \\
    Masked CC Generated 100\%   & visual\_bert.pretrained.cc.full\_coco\_generated     \\
    Masked CC Generated 50\%   & visual\_bert.pretrained.cc.half\_coco\_generated     \\
    Masked CC Generated 10\%   & visual\_bert.pretrained.cc.small\_coco\_generated     \\
    \bottomrule
  \end{tabular}
  \label{tab:models}
\end{table}

\begin{table}[H]
 \caption{scores of top 3 pretrained models}
  \centering
  \begin{tabular}{ll}
    \toprule
    Pretrained model          & AUROC \\
    \midrule
    visual\_bert.pretrained.cc.full  & 0.7260\\
    visual\_bert.pretrained.cc.small\_fifty\_pc  & 0.7396\\
    visual\_bert.pretrained.coco.full  & 0.7355\\
    \bottomrule
  \end{tabular}
  \label{tab:scores} 
\end{table}

\subsection{Model Enhancement}
\label{subsec:enhancement}
UNITER outperformed other SOTA models such as VisualBert, VilBERT, LXMERT. It introduce Masked Region Modeling(MRM) and Masked Language Modeling(MLM) into single-stream model, which randomly mask out the visual features and the input words with a probability of 15\% respectively. Similar to  VisualBert, it use region features as visual feature, but implement region mask on region encodes. This inspired me to implement MRM and/or MLM to pre-trained models of VisualBERT which listed in table \ref{tab:models}. The result shows MRM improves the performance of some pre-trained models, but MLM not. The scores of top 3 improved models are shown in table \ref{tab:MRMscores}.

After carefully analysis, I have selected top 4 pre-trained models in my experiments as base models, which is the prototype in fig\ref{fig:process}, they are:
\begin{itemize}
   \item  visual\_bert.pretrained.cc.full+MRM
   \item  visual\_bert.pretrained.coco.full+MRM
   \item  visual\_bert.finetuned.hateful\_memes.direct+MRM
   \item  visual\_bert.pretrained.cc.small\_fifty\_pc
\end{itemize}

\begin{table}[H]
 \caption{scores of top 3 models with MRM}
  \centering
  \begin{tabular}{ll}
    \toprule
    Pretrained model          & AUROC \\
    \midrule
    visual\_bert.pretrained.cc.full+MRM  & 0.7400\\
    visual\_bert.pretrained.coco.full+MRM  & 0.7376\\
    visual\_bert.finetuned.hateful\_memes.direct+MRM &  0.7386\\
    \bottomrule
  \end{tabular}
  \label{tab:MRMscores}
\end{table}

\subsection{Semi-supervised Learning}
The proportion of "3-tuple" memes is about 40\% in dev sets or test sets, they are difficult samples for any multimodal models. The implementation of rule 1 can achieve an accuracy of about 98\% for “3-tuple” in dev set. Assuming that the same rule is met on test set, the labels of these “3-tuple” samples will almost be certain. This means they are not “unseen” for me, if I add these samples to training set, the classification accuracy will be greatly improved. So I extracted the “3-tuple” samples from test seen and test unseen, labeled them by rule 1 before merge them into training set. The experimental results verify my conjecture, both AUROC and accuracy have been significantly improved (over 10.0\%). which proves that this method also have a positive impact on the classification accuracy of independence samples.

\subsection{Adjustment Based on Rule 2}
\label{subsec:rules}
"2-tuple" memes are difficult samples for multimodal model, and the percentage or them is about 10\% in dev sets or test sets. For “2-tuple”, the classification probabilities adjusted by rule 2 can improve the accuracy. In addition, compare with adjust the result after models stacking, adjust the result of every base model before implement stacking is more effective, with improvement of AUROC of over 1.5\% and over 3.2\% respectively. 
\subsection{Models Stacking}
\label{subsec:stacking}
To reduce over-fitting and obtain as much effective information as possible from limited data, I applied K-fold in training process. K-fold is a model validation technique that provides train/validation indices to split data in train and validation sets. It will split dataset into $k$ consecutive folds. Each fold is then used a validation set once while the other $k-1$ remaining folds form the training set. 

I set $k=5$ in my experiment, each model applied 5-fold technique can produce 5 different models. By applying 5-fold technique to 4 models mentioned in section \ref{subsec:enhancement}, 20 different models are obtained, that is, 20 classification results can be obtained. The simplest way to generate the final probability from 20 results is equal weight, that is, average 20 probabilities as the final probability for each test id. When the final probability is not less than 0.5, the label is set to 1, vice versa. Equal weight is simple but effective, which outperforms other stacking method such as logistic regression, decision tree and MLP in my attempts. 
\section{Discussions}
In this challenge,I implemented MRM to VisualBert to enhance its effectiveness in Hateful Memes, and applied several common technologies such as K-fold, model stacking, semi-supervised learning, significantly improved the AUROC and accuracy of classification. The most important of my work is the combination of rules extracted from data set with multimodal framework, which improved both the accuracy and the AUROC more than 13\%. The features of the dataset are too conspicuous, this makes it possible for me to extract rules from data, but these rules may not be so practical in real world scenarios. On the other hand, my work shows that the performance of multimodal models on difficult samples is not so good. So the attempts to improve the multimodal framework in the future should  focus on it.
\bibliographystyle{unsrt}  
\bibliography{references}

\begin{thebibliography}{10}

\bibitem{kiela2020hateful}
Douwe Kiela, Hamed Firooz, Aravind Mohan, Vedanuj Goswami, Amanpreet Singh,
  Pratik Ringshia, and Davide Testuggine.
\newblock The hateful memes challenge: Detecting hate speech in multimodal
  memes.
\newblock {\em arXiv preprint arXiv:2005.04790}, 2020.

\bibitem{devlin2018bert}
Jacob Devlin, Ming-Wei Chang, Kenton Lee, and Kristina Toutanova.
\newblock Bert: Pre-training of deep bidirectional transformers for language
  understanding.
\newblock {\em arXiv preprint arXiv:1810.04805}, 2018.

\bibitem{liu2019roberta}
Yinhan Liu, Myle Ott, Naman Goyal, Jingfei Du, Mandar Joshi, Danqi Chen, Omer
  Levy, Mike Lewis, Luke Zettlemoyer, and Veselin Stoyanov.
\newblock Roberta: A robustly optimized bert pretraining approach.
\newblock {\em arXiv preprint arXiv:1907.11692}, 2019.

\bibitem{lan2019albert}
Zhenzhong Lan, Mingda Chen, Sebastian Goodman, Kevin Gimpel, Piyush Sharma, and
  Radu Soricut.
\newblock Albert: A lite bert for self-supervised learning of language
  representations.
\newblock {\em arXiv preprint arXiv:1909.11942}, 2019.

\bibitem{sun2019videobert}
Chen Sun, Austin Myers, Carl Vondrick, Kevin Murphy, and Cordelia Schmid.
\newblock Videobert: A joint model for video and language representation
  learning.
\newblock In {\em Proceedings of the IEEE International Conference on Computer
  Vision}, pages 7464--7473, 2019.

\bibitem{sun2019learning}
Chen Sun, Fabien Baradel, Kevin Murphy, and Cordelia Schmid.
\newblock Learning video representations using contrastive bidirectional
  transformer.
\newblock {\em arXiv preprint arXiv:1906.05743}, 2019.

\bibitem{lu2019vilbert}
Jiasen Lu, Dhruv Batra, Devi Parikh, and Stefan Lee.
\newblock Vilbert: Pretraining task-agnostic visiolinguistic representations
  for vision-and-language tasks.
\newblock In {\em Advances in Neural Information Processing Systems}, pages
  13--23, 2019.

\bibitem{tan2019lxmert}
Hao Tan and Mohit Bansal.
\newblock Lxmert: Learning cross-modality encoder representations from
  transformers.
\newblock {\em arXiv preprint arXiv:1908.07490}, 2019.

\bibitem{alberti2019fusion}
Chris Alberti, Jeffrey Ling, Michael Collins, and David Reitter.
\newblock Fusion of detected objects in text for visual question answering.
\newblock {\em arXiv preprint arXiv:1908.05054}, 2019.

\bibitem{li2019visualbert}
Liunian~Harold Li, Mark Yatskar, Da~Yin, Cho-Jui Hsieh, and Kai-Wei Chang.
\newblock Visualbert: A simple and performant baseline for vision and language.
\newblock {\em arXiv preprint arXiv:1908.03557}, 2019.

\bibitem{li2020unicoder}
Gen Li, Nan Duan, Yuejian Fang, Ming Gong, Daxin Jiang, and Ming Zhou.
\newblock Unicoder-vl: A universal encoder for vision and language by
  cross-modal pre-training.
\newblock In {\em AAAI}, pages 11336--11344, 2020.

\bibitem{su2019vl}
Weijie Su, Xizhou Zhu, Yue Cao, Bin Li, Lewei Lu, Furu Wei, and Jifeng Dai.
\newblock Vl-bert: Pre-training of generic visual-linguistic representations.
\newblock {\em arXiv preprint arXiv:1908.08530}, 2019.

\bibitem{zhou2020unified}
Luowei Zhou, Hamid Palangi, Lei Zhang, Houdong Hu, Jason~J Corso, and Jianfeng
  Gao.
\newblock Unified vision-language pre-training for image captioning and vqa.
\newblock In {\em AAAI}, pages 13041--13049, 2020.

\bibitem{antol2015vqa}
Stanislaw Antol, Aishwarya Agrawal, Jiasen Lu, Margaret Mitchell, Dhruv Batra,
  C~Lawrence~Zitnick, and Devi Parikh.
\newblock Vqa: Visual question answering.
\newblock In {\em Proceedings of the IEEE international conference on computer
  vision}, pages 2425--2433, 2015.

\bibitem{chen2020uniter}
Yen-Chun Chen, Linjie Li, Licheng Yu, Ahmed El~Kholy, Faisal Ahmed, Zhe Gan,
  Yu~Cheng, and Jingjing Liu.
\newblock Uniter: Universal image-text representation learning.
\newblock In {\em European Conference on Computer Vision}, pages 104--120.
  Springer, 2020.

\bibitem{lu202012}
Jiasen Lu, Vedanuj Goswami, Marcus Rohrbach, Devi Parikh, and Stefan Lee.
\newblock 12-in-1: Multi-task vision and language representation learning.
\newblock In {\em Proceedings of the IEEE/CVF Conference on Computer Vision and
  Pattern Recognition}, pages 10437--10446, 2020.

\bibitem{singh2020mmf}
Amanpreet Singh, Vedanuj Goswami, Vivek Natarajan, Yu~Jiang, Xinlei Chen, Meet
  Shah, Marcus Rohrbach, Dhruv Batra, and Devi Parikh.
\newblock Mmf: A multimodal framework for vision and language research.
\newblock \url{https://github.com/facebookresearch/mmf}, 2020.

\end{thebibliography}

\end{document}